\documentclass[manuscript, screen]{acmart}

\AtBeginDocument{%
  \providecommand\BibTeX{{%
    \normalfont B\kern-0.5em{\scshape i\kern-0.25em b}\kern-0.8em\TeX}}}

\usepackage{multirow}
\usepackage{array}

\usepackage{pifont}
\newcommand{\xmark}{\ding{55}}
\newcommand{\cmark}{\ding{51}}
\usepackage{makecell}
\usepackage[edges]{forest}

\usepackage{xcolor, color, soul}
\sethlcolor{yellow}

\begin{document}

\title{Local Interpretations for Explainable Natural Language Processing: A Survey}


\author{Siwen Luo}
\authornote{Both authors contributed equally to this research.}
\email{siwen.luo@uwa.edu.au}
\affiliation{
  \institution{The University of Western Australia}
  \streetaddress{35 Stirling Hwy, Crawley}
  \city{Perth}
  \state{WA}
  \country{Australia}
  \postcode{6009}
}

\author{Hamish Ivison}
\authornotemark[1]
\email{hamishiv@cs.washington.edu}
\affiliation{
  \institution{University of Washington}
  \streetaddress{3800 E Stevens Way NE}
  \city{Seattle}
  \state{WA}
  \country{United States}
  \postcode{98195-2355}
}

\author{Soyeon Caren Han}
\authornote{Corresponding Author}
\email{caren.han@unimelb.edu.au}
\affiliation{%
  \institution{The University of Melbourne}
  \streetaddress{700 Swanston Street, Carlton}
  \city{Melbourne}
  \state{VIC}
  \country{Australia}
  \postcode{3010}
}

\author{Josiah Poon}
\email{josiah.poon@sydney.edu.au}
\affiliation{%
  \institution{The University of Sydney}
  \streetaddress{1 Cleveland St, Darlington}
  \city{Sydney}
  \state{NSW}
  \country{Australia}
  \postcode{2008}
}

\renewcommand{\shortauthors}{Luo and Ivison, et al.}

\begin{abstract}
  As the use of deep learning techniques has grown across various fields over the past decade, complaints about the opaqueness of the black-box models have increased, resulting in an increased focus on transparency in deep learning models. This work investigates various methods to improve the interpretability of deep neural networks for Natural Language Processing (NLP) tasks, including machine translation and sentiment analysis. We provide a comprehensive discussion on the definition of the term \textit{interpretability} and its various aspects at the beginning of this work. The methods collected and summarised in this survey are only associated with local interpretation and are specifically divided into three categories: 1) interpreting the model's predictions through related input features; 2) interpreting through natural language explanation; 3) probing the hidden states of models and word representations.
\end{abstract}



\maketitle

\section{Introduction}
As a result of the explosive development of deep learning techniques over the past decade, the performance of deep neural networks (DNN) has significantly improved across various tasks. DNN has been broadly applied in different fields, including business, healthcare, and justice. For example, in healthcare, artificial intelligence startups have raised \$864 million in the second quarter of 2019, with higher amounts expected in the future as reported by the TDC Group\footnote{\url{https://www.thedoctors.com/articles/the-algorithm-will-see-you-now-how-ais-healthcare-potential-outweighs-its-risk/}}. However, while deep learning models have brought many foreseeable benefits to both patients and medical practitioners, such as enhanced image scanning and segmentation, the inability of these models to provide explanations for their predictions is still a severe risk, limiting their application and utility.

Before demonstrating the importance of the interpretability of deep learning models, it is essential to illustrate the opaqueness of DNNs compared to other interpretable machine learning models. Neural networks roughly mimic the hierarchical structures of neurons in the human brain to process information among hierarchical layers. Each neuron receives the information from its predecessors and passes the outputs to its successors, eventually resulting in a final prediction~\citep{mueller2019deep}. DNNs are neural networks with a large number of layers, meaning they contain up to billions of parameters. Compared to interpretable machine learning models such as linear regressions, where the few parameters in the model can be extracted as the explanation to illustrate influential features in prediction, or the decision trees, where a model's prediction process can be easily understood by following the decision rules, the complex and huge computations done by DNNs are hard to comprehend both for experts and non-experts alike. In addition, the representations used and constructed by DNNs are often complex and incredibly difficult to tie back to a set of observable variables in image and natural language processing tasks. As such, vanilla DNNs are often regarded as opaque `black-box' models that have neither interpretable architectures nor clear features for interpretation of the model outputs. 

However, why should we want interpretable DNNs? One fundamental reason is that while the recent application of deep learning techniques to various tasks has resulted in high levels of performance and accuracy, these techniques still need improvement. As such, when applying these models to critical tasks where prediction results can cause significant real-world impacts, they are not guaranteed to provide faultless predictions. Furthermore, given any decision-making system, it is natural to demand explanations for the decisions provided. For example, the European Parliament adopted the General Data Protection Regulation (GDPR) in May 2018 to clarify the right of explanation for all individuals to obtain “meaningful explanations of the logic involved” for automated decision-making procedures~\citep{Guidotti2018}. As such, it is legally and ethically crucial for the application of DNNs to develop and design ways for these networks to provide explanations for their predictions. In addition, explanations of predictions would help specialists verify their correctness, allowing them to judge if a model is making the right predictions for the right reasons. As such, increasing interpretability is vital for expanding the applicability and correctness of DNNs.

In the past few years, several works have been proposed to improve the interpretability of DNNs. In this survey paper, we focus on \textit{local} interpretable methods proposed for natural language processing tasks. As described in the following sections, we define \textit{local methods} as those which provide explanations only for specific decisions made by the model - that is, methods that provide explanations for single instances rather than aiming to provide general descriptions of the model's decision-making process. We explore several recent local interpretation methods/techniques in Natural Language Processing (NLP), which aims to support the normal users with no machine/deep learning expertise\footnote{Note that there are several local interpretation methods, such as counterfactuals, example-based approaches, have not been included in this article since only a few initial NLP research tasks have been conducted with these example-based approaches.}:

\begin{itemize}
    \item \textbf{Feature importance methods}, which work by determining and extracting the most important elements of an input instance.
    \item \textbf{Natural language explanation}, in which models generate text explanations for a given prediction.
    \item \textbf{Probing}, in which model's internal states are examined when given certain inputs.
\end{itemize}

\subsection{Definitions of Interpretability}
While there has been much study of the interpretability of DNNs, there are no unified definitions for the term \textit{interpretabilty}, with different researchers defining it from different perspectives. We summarise the key aspects of interpretability used by these researchers below.

\subsubsection{Explainability vs Interpretability}\label{sec: Explainability vs Interpretability}

The terms \textit{interpretability} and \textit{explainability} are often used synonymously across the field of explainable AI \citep{chakraborty2017, adadi2018peeking}, with both terms being used to refer to the ability of a system to justify or explain the reasoning behind its decisions\footnote{For example, \citet{stadelmaier-pado-2019-modeling, stahlberg-etal-2018-operation, liu-etal-2019-towards-explainable, wang-etal-2019-multi-granular} primarily use \textit{explainability} or \textit{explainable}, while \citet{serrano-smith-2019-attention, lime, esnli, tutek-snajder-2018-iterative} primarily use \textit{interpretable} or \textit{interpretability}.}. Overall, the machine learning community tends to use the term \textit{interpretability}, while the HCI community tends to use the term \textit{explainability} \citep{adadi2018peeking}. Recent work has suggested more formal definitions of these terms \citep{chakraborty2017, rigscience, Guidotti2018}. Following \citet{rigscience}, we define \textit{interpretability} as `\textit{the ability [of a model] to explain or to present [its predictions] in understandable terms to a human}.' We take \textit{explainability} to be synonymous with interpretability unless otherwise stated, reflecting its general usage within the field.

\subsubsection{Local and Global Interpretability}

An essential distinction in interpretable machine learning is between \textit{local} and \textit{global} interpretability. Following \citet{Guidotti2018} and \citet{rigscience}, we take \textit{local interpretability} to be `\textit{the situation in which it is possible to understand only the reasons for a specific decision}' \citep{Guidotti2018}. That is, a locally interpretable model is a model that can give explanations for specific predictions and inputs. We take \textit{global interpretability} to be the situation in which it is possible to understand `\textit{the whole logic of a model and follow the entire reasoning leading to all the different possible outcomes}' \citep{Guidotti2018}. A classic example of a globally interpretable model is a decision tree, in which the general behaviour of the model may be easily understood by examining the decision nodes that make up the tree. As understanding the whole logic of a model often requires the use of specific models or significant changes to an existing model, in this paper, we focus on local interpretation methods, as these tend to be more generally applicable to existing and future NLP models.

\subsubsection{Post-hoc vs In-built Interpretations}
Another important distinction is whether an interpretability method is applied to a model after the fact or integrated into the internals of a model. The former is referred as a \textit{post-hoc} interpretation method \cite{molnar2019}, while the latter is an \textit{in-built} interpretation method. As \textit{Post-hoc} methods are applied to model the fact, they generally do not impact the model's performance. Some post-hoc methods do not require any access to the internals of the model being explained and so are \textit{model-agnostic}. An example of a typical post-hoc interpretable method is LIME \cite{ribeiro2016should}, which generates the local interpretation for one instance by permuting the original inputs of an underlying black-box model. In contrast to post-hoc interpretations, \textit{in-built interpretations} are closely integrated into the model itself. The interpretation may come from the transparency of the model, where the workings of the model itself are clear and easy to understand (for example, a decision tree), or may come from an interpretation generated by the model in an opaque manner (for example, a model that generates a text explanation during its prediction process). In this survey, we will examine both methods.

\subsection{Paper layout}
Before examining interpretability methods, we first discuss different aspects of interpretability in Section 2. In Section 3 we summarize and categorize three main interpretable methods in NLP, including 1) improving a model's interpretability by identifying the important input features; 2) explaining a model's predictions by generating direct natural language explanations; 3) probing the internal state and mechanisms of a model. We also provide a quick summary of datasets that are commonly used for the study of each method. In Section 4, we summarise several primary methods to evaluate the interpretability of each method discussed in Section 3. We finally discussed the limitations of current interpretable methods in NLP in Section 5 and the possible future trend of interpretability development at the end.

\section{Aspects of Interpretability}

\subsection{Interpretability requirements}

Before discussing the various aspects of interpretability, it is also essential to consider what problems require interpretable solutions and what interpretable models best fit these problems. Following \cite{rigscience}, we suggest that anyone looking to build interpretable models first determine the following four points:
\begin{enumerate}
    \item \textit{Do you need an explanation for a specific instance or understand how a model works?} In the former case, local interpretation methods will likely prove more suitable, while in the latter global interpretation methods will be required.
    \item \textit{How much time does/will a user have to understand the explanation?} This, along with the point below, is an important concern for the usability of an interpretation method. Certain methods lend themselves to quick, intuitive understanding, while others require some more effort and time to comprehend.
    \item \textit{What background and expertise will the users of your interpretable model have?} As mentioned, this is an important usability concern. For example, regression weights have classically been considered `interpretable', but require a user to have some understanding of regression beforehand. In contrast, decision trees (when rendered in a tree structure) are often understandable even to non-experts.
    \item \textit{What aspects or parts of the problem do you want to explain?} It is important to consider what can and cannot be explained by your model, and prioritise accordingly. For example, explaining all potential judgements a self-driving car could make in any situation is infeasible, but restricting explanations to certain systems or situations allows easier measuring and assurance of interpretation quality.
\end{enumerate}

These points allow categorisation of interpretability-related problems, and thus clearer understanding of what is required from an interpretable system and suitable interpretation methods for the problem itself.

\subsection{Dimensions of Interpretability}

`Interpretability' is not a simple binary or monolithic concept, but rather one that can be measured along multiple dimensions. Different aspects of interpretability have been identified across the literature, which we condense and summarise into four key aspects: \textit{faithfulness}, \textit{stability}, \textit{comprehensibility}, and \textit{trustworthiness}.

\subsubsection{Faithfulness}
Faithfulness measures how well an interpretation method reflects the decision-making process used by the underlying model. For example, an image heatmap that highlights parts of the image not genuinely used by the model would be unfaithful, while highlighting the parts genuinely used by the model would be more faithful. Traditionally, this has been more a concern for post-hoc methods such as LIME \cite{ribeiro2016should} and SHAP \cite{lundberg2017unified}. However, more recent work has called into question the faithfulness of in-built interpretability methods such as attention weight examination \cite{wiegreffe-pinter-2019-attention, jacovi-goldberg-2020-towards, jain-wallace-2019-attention}. Faithfulness is essential for claims that an interpretation method accurately reflects a model's process to reach a judgement. Explanations provided by an unfaithful method may hide existing biases that the underlying model uses for judgements, potentially engendering unwarranted trust or belief in these predictions \cite{jacovi-goldberg-2020-towards}. Related is the notion of \textit{fidelity} as defined in \citet{molnar2019}: the extent of how well an interpretable method can approximate the performance of a black-box model. Underlying this definition is the assumption that a method that better approximates a black box also must use a similar reasoning process to that underlying model\footnote{This is stated as `the model assumption' in \citet{jacovi-goldberg-2020-towards}.}. As such, this definition of fidelity is a more specific form of faithfulness as applied to interpretability methods that construct models approximating an underlying black-box model, such as LIME \citep{lime}.

\subsubsection{Stability}
An interpretation method is stable if it provides similar explanations for similar inputs \citep{molnar2019} unless the difference between the inputs is highly important for the task at hand. For example, an explanation produced by natural language generation (NLG) would be stable if minor differences in the input resulted in similar text explanations and would be unstable if the slight differences resulted in wildly different explanations. Stability is a generally desirable trait important for research \citep{yu2013} and is required for a model to be trustworthy \citep{Murdoch22071}. In addition, the stability of human explanations for a particular task should be considered, i.e. if explanations written by humans differ significantly from each other, it is unreasonable to expect a model trained on such explanations to do any better. This is especially important for highly free-form interpretation methods such as natural language explanations.

\subsubsection{Comprehensibility}
An interpretation is considered \textit{comprehensible} if it's understandable to an end-user. In order for an explanation to be useful at all, it must be understandable to some degree. However, this is subjective: there is no global common standard for `understandability'. In addition, the background of the end-user matters: a medical professional will be able to understand an explanation with scientific medical terms far better than a layperson. Nevertheless, there are still several general ways to rate the interpretability of an explanation: examining its size (how much a user must process when `reading' the explanation), examining how well a human can predict a model's prediction given just the explanation, and examining the understandability of individual features of the explanation \cite{molnar2019}. For example, a sparse linear model with only a few non-zero weights has far fewer components for a user to consider, and so would be more comprehensible than a linear model with hundreds of weights. Furthermore, comprehensibility is related to the concept of \textit{transparency} \citep{Lipton2018}, which refers to how well a person can understand the mechanism by which a model works. Transparency can be achieved in several ways: through being able to simulate the model in your mind (for example, a linear regression with few weights), or having deep knowledge of the underlying algorithm used by the model (for example, proving some property of any solution an algorithm will produce). Models with greater degrees of transparency are thus also more comprehensible than non-transparent models.

\subsubsection{Trustworthiness}
An interpretation is considered \textit{comprehensible} if it is understandable to an end-user. In order for an explanation to be helpful at all, it must be understandable to some degree. However, this is subjective. In other words, there is no global shared standard for `understandability'. In addition, the background of the end-user matters: a medical professional will be able to understand an explanation with scientific medical terms far better than a layperson. Nevertheless, there are still several general ways to rate the interpretability of an explanation: examining its size (how much a user must process when `reading' the explanation), examining how well a human can predict a model's prediction given just the explanation, and examining the understandability of individual features of the explanation \cite{molnar2019}. For example, a sparse linear model with only a few non-zero weights has far fewer components for a user to consider and would be more comprehensible than a linear model with hundreds of weights. Furthermore, comprehensibility is related to the concept of \textit{transparency} \citep{Lipton2018}, which refers to how well a person can understand the mechanism by which a model works. There are several ways to achieve \textit{transparency}: through being able to simulate the model in your mind (for example, a linear regression with few weights) or having deep knowledge of the underlying algorithm used by the model (for example, proving some property of any solution an algorithm will produce). Models with greater transparency degrees are thus more comprehensible than non-transparent models.

\begin{figure*}
    \centering
    \includegraphics[width=\textwidth]{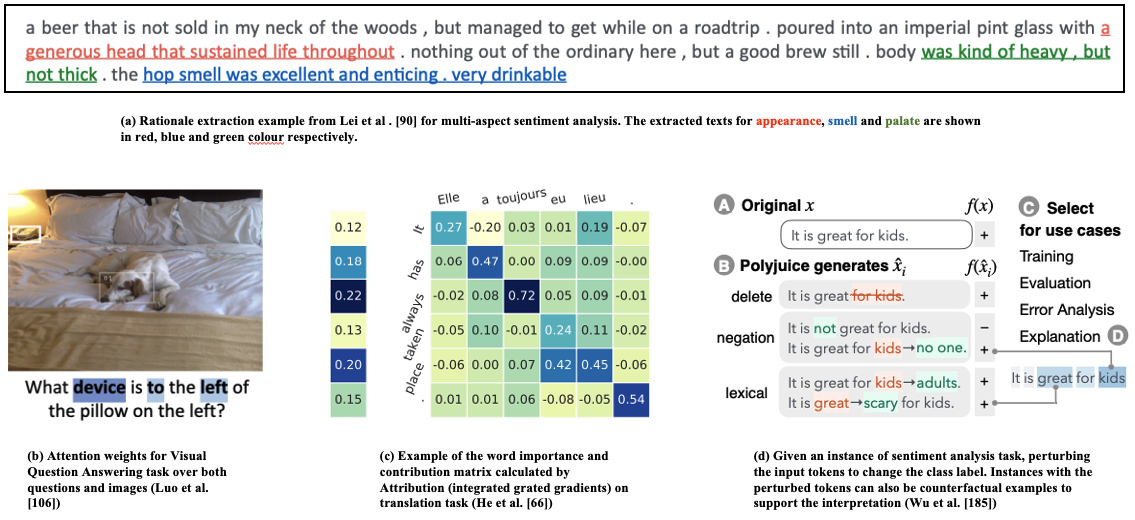}
    \caption{Sample visualizations of identified important features from the inputs detected by four different methods. (a): Rationale Extraction on sentiment analysis task; (b) Attention Weights on Visual Question Answering task: (c) Word importance from Attribution methods on machine translation task; (d) Input perturbation on sentiment analysis task and the expansion of counterfactual explanation.}
    \label{fig:visual}
\end{figure*}

\section{Interpretable Methods}

\subsection{Feature Importance}
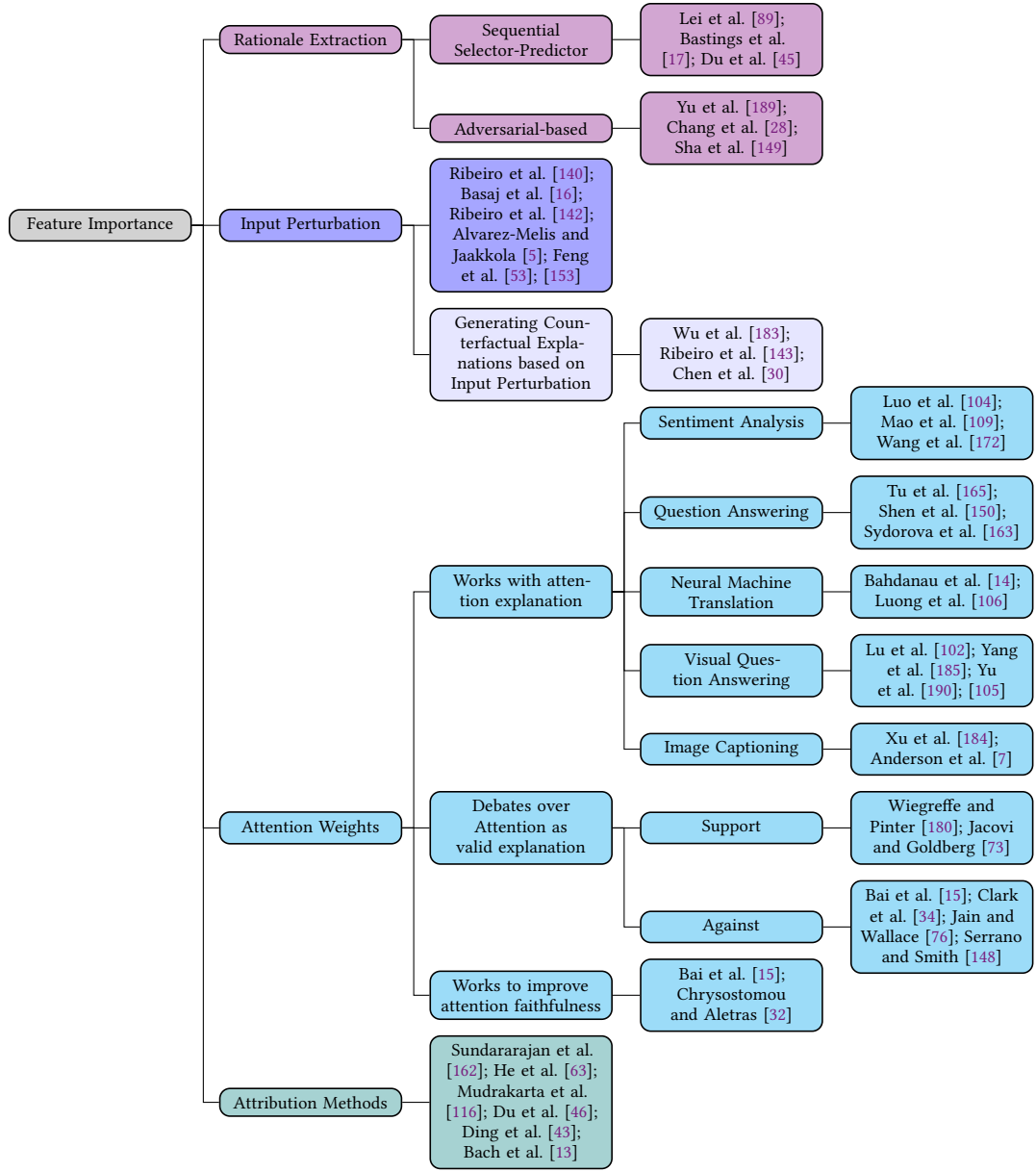
\begin{figure*}
\begin{forest}
  for tree={
    forked edges,
    font=\footnotesize,
    draw,
    grow' = 0,
    semithick,
    rounded corners,
    text width = 2.3cm,
    s sep = 6pt,
    node options = {align = center},
    calign=child edge, 
    calign child=(n_children()+1)/2
  }
  [Feature Importance, fill=gray!35
    [Rationale Extraction, for tree={fill=violet!35}
        [Sequential Selector-Predictor
            [\citet{lei2016rationalizing};
            \citet{bastings2019interpretable};
            \citet{Du2019LearningCD}]
        ]
        [Adversarial-based
            [\citet{yu2019rethinking};
            \citet{chang2019game};
            \citet{sha2021learning}
            ]
        ]
    ]
    [Input Perturbation, for tree={fill=blue!35}
        [\citet{ribeiro2016should};
        \citet{Basaj2018HowMS};
        \citet{ribeiro2018anchors};
        \citet{alvarez2017causal};
        \citet{feng2018pathologies};
        \cite{slack2020fooling}
        ]
        [Generating Counterfactual Explanations based on Input Perturbation, for tree={fill=blue!10}
            [\citet{wu2021polyjuice};
            \citet{ribeiro2020beyond};
            \citet{chen2021kace}
            ]
        ]
    ]
    [Attention Weights, for tree={fill=cyan!35}
        [Works with attention explanation
            [Sentiment Analysis
                [\citet{luo2018beyond};
                \citet{mao2019aspect};
                \citet{wang2018aspect}
                ]
            ]
            [Question Answering
                [\citet{tu2020select};
                \citet{shen2018knowledge};
                \citet{sydorova2019interpretable}
                ]
            ]
            [Neural Machine Translation
                [\citet{bahdanau2014neural};
                \citet{luong2015effective}
                ]
            ]
            [Visual Question Answering
                [\citet{lu2016hierarchical};
                \citet{yang2016stacked};
                \citet{yu2019deep};
                \cite{luo2020rexup}
                ]
            ]
            [Image Captioning
                [\citet{xu2015show};
                \citet{anderson2018bottom}
                ]
            ]
        ]
        [Debates over Attention as valid explanation
            [Support
                [\citet{wiegreffe2019attention};
                \citet{jacovi2020towards}
                ]
            ]
            [Against
                [\citet{bai2021attentions};
                \citet{clark2019does};
                \citet{jain2019attention};
                \citet{serrano2019attention}
                ]
            ]
        ]
        [Works to improve attention faithfulness
            [\citet{bai2021attentions};
            \citet{chrysostomou2021improving}
            ]
        ]
    ]
    [Attribution Methods, for tree={fill=teal!35}
        [\citet{sundararajan2017axiomatic};
        \citet{he2019towards};
        \citet{mudrakarta2018did};
        \citet{du2019attribution};
        \citet{ding2017visualizing};
        \citet{bach2015pixel}
        ]
    ]
  ]
\end{forest}
\caption{Typology of local interpretable methods by identifying the important features from inputs.}
\label{fig:typo-feature}
\end{figure*}

Identifying the important input features that significantly impact a model's prediction results is a straightforward method of improving a model's local interpretability, directly linking model outputs to inputs. Important features can be, for example, words for text-based tasks or image regions for image-based tasks. This paper focuses on the four main different methods of extracting important features as the interpretation for the model's outputs: rationale extraction, input perturbation, attribution methods and attention weight extraction. We conclude the typology of feature importance methods in Figure~\ref{fig:typo-feature} and present the sample visualisations of extracted features from inputs in Figure~\ref{fig:visual}.

\subsubsection{Rationale Extraction}
Rationale extractions are usually used as the local interpretable method for NLP tasks of sentiment analysis and document classification. Rationales are short and coherent phrases from the original textual inputs and represent the critical textual features that contribute most to the output prediction. These identified textual features work as the local explanation that interprets the information the model primarily pays attention to when making the prediction decision for a particular textual input. The good rationales valid for the explanation should lead to the same prediction results as the original textual inputs. As this work area developed, researchers also made extra efforts to extract coherent and consecutive rationales to use them as more readable and comprehensive explanations. 

The rationale extraction methods can be mainly divided into two streams: 1). a sequential selector-predictor stacked model, where the selector first selects the rationales from the original textual inputs and then pass to the predictor for the prediction result; 2). the adversarial-based model that involves the parallel models to calibrate the rationales extracted by the selector. In this paper, we summarise several iconic and milestone works of rationale extractions for each stream.

For the selector-predictor stream, \citet{lei2016rationalizing} is one of the first works for the rationale extraction in NLP tasks. The selector process first generates a binary vector of 0 and 1 through a Bernoulli distribution conditioned on the original textual inputs. This binary vector will then be multiplied over the original inputs where 1 indicates the selection of input words as rationales and 0 indicates the non-selection, resulting in a sparse input representation that indicates which textual tokens are selected as rationales and which tokens are not. The predictor will then process based on such information. Since the selected rationales are represented with non-differentiable discrete values, the REINFORCE algorithm \cite{williams1992simple} was applied for optimization to update the binary vectors for the eventually accurate rational selection. \citet{lei2016rationalizing} performed rationale extraction for a sentiment analysis task with the training data that has no pre-annotated rationales to guide the learning process. The training loss is calculated through the difference between a ground truth sentiment vector and a predicted sentiment vector generated from extracted rationales selected by the selector model. Such selector-predictor structure is designed to mainly boost the interpretability faithfulness, i.e. selecting valid rationales that can predict the accurate output as the original textual inputs. To increase the readiness of the explanation, \citet{lei2016rationalizing} used two different regularizers over the loss function to force rationales to be consecutive words (readable phrases) and limit the number of selected rationales (i.e. selected words/phrases). \citet{bastings2019interpretable} followed the same selector-predictor structure as \citet{lei2016rationalizing}. The main difference is that they used rectified Kumaraswamy distribution~\cite{kumaraswamy1980generalized} instead of Bernoulli distribution to generate the rationale selection vector, i.e. the binary vector of 0 and 1 to be masked over textual inputs. Kumaraswamy distribution allows the gradient estimation for optimization, so there is no need for the REINFORCE algorithm to do the optimization. To boost the short and coherent rationales for better readability and comprehensibility, \citet{bastings2019interpretable} also applied a relaxed form of $L_0$ regularization \cite{louizos2018learning} and the Lagrangian relaxation to encourage adjacent words selected or not selected together. Different from the above methods, where rationale extraction is wrapped in an end-to-end model and has not used annotated rationales during the training of rationale selection, \citet{Du2019LearningCD} uses rationales annotated by external experts as guidance during the training of rationale selector to generate the local explanations (short and coherent rationales) that are consistent with these external human-annotated rationales.

For the stream of adversarial-based models, a third module is usually added in addition to the selector-predictor stacks, functioning as a guide to boost the faithfulness of rationales and improve the comprehensibility of interpretation. For example, to boost the faithfulness of extracted rationales, \citet{yu2019rethinking} inserted the target labels of sentiment analysis as additional inputs into the rationale selector to boost its participation in prediction. Additionally, to improve the comprehensibility that prevents the rationale selector from selecting meaningless small snippets, this work added a third element: a complement predictor. This additional module predicts the labels for original textual inputs based on non-rationale words. The complement predictor and the generator work much like the discriminative and generative networks in generative adversarial networks (GANs) \cite{goodfellow2014generative}: the rationale selector aims to extract as many prediction-relevant words as possible as rationales to avoid the complement predictor from being able to predict the actual textual label. Similar to \citet{yu2019rethinking}, \citet{chang2019game} also involved a third module where the target labels of the original inputs are used as additional inputs, but with the addition that these target labels can be incorrect. This work also proposed a counterfactual rational generator to extract relevant rationales that cause false predictions. A discriminator is then applied to discriminate between the actual and counterfactual rationale generator. Recent work such as \cite{sha2021learning} reduces the complexity of using three modules but constructs a guider model that operates over the original textual inputs for prediction and the rationale selector model in the adversarial-based architecture to encourage the final prediction vectors from two separate models to be close to each other, and thus achieve the faithfulness of extracted rationales. Also, to achieve better comprehensibility, \cite{sha2021learning} proposed language models as a regularizer, which significantly contributes to the better fluency of the extracted rationale by selecting consecutive tokens that describe the rationale well. 

In general, using extracted rationales from original textual inputs as the models' local interpretations focuses on the faithfulness and comprehensibility of interpretations. While trying to select rationales that can well represent the complete inputs in terms of accurate prediction results, extracting short and consecutive sub-phrases is also the key objective of the current rationale extraction works. Such fluent and consecutive sub-phrases (i.e. the well-extracted rationales) make this rationales extraction a friendly, interpretable method that provides readable and understandable explanations to non-expert users without NLP-related knowledge.

\subsubsection{Input Perturbation}


Another method for identifying important features of textual inputs is input perturbation. For this method, a word (or a few words) of the original input is modified or removed (i.e. `perturbed'), and the resulting performance change is measured. The more significant the model's performance drop, the more critical these words are to the model and therefore are regarded as important features. Input perturbation is usually model-agnostic, which does not influence the original model's architecture. The main difference among the proposed input perturbation methods lies in how to perturb the tokens or phrases from original inputs into the new instances.

\citet{ribeiro2016should} proposed a local interpretable model-agnostic explanations (LIME) model that can be used as an interpretable method for any black-box model. The main idea of LIME is the approximation of a black-box model with a transparent model using variants of original inputs. For natural language processing tasks such as text classification, words of original textual inputs are randomly selected and removed from the inputs, using a binary representation to mark the inclusion of words. \citet{Basaj2018HowMS} applied LIME to a QA task for identifying the important words in a question, where the words in the questions are considered to be features, while the associated context (i.e. text containing the answer to the given question) was held constant. The results indicate that in QA tasks, the complete sentence of question plays a minor role, and just a small amount of question words are sufficient for correct answer prediction.

\citet{ribeiro2018anchors} argued that the important features identified by \citet{ribeiro2016should} are based on word-level (single token) instead of phrase-level (consecutive tokens) features. Word-level features relate to only one instance and cannot provide general explanations, which makes it difficult to extend such explanations to unseen instances. For example, in sentiment analysis, `\textit{not}' in `\textit{The movie is not good}' is a contributing feature for negative sentiment but is not a contributing feature for positive sentiment in `\textit{The weather is not bad}'. The single token `\textit{not}' is insufficient as a general explanation for unseen instances as it will lead to different meanings when combined with different words. Thus, \citet{ribeiro2018anchors} emphasized the phrase-level features for more comprehensive local interpretations and proposed a rule-based method for identifying critical features for predictions. Their proposed algorithm iteratively selects predicates from inputs as key tokens while replacing the rest of the tokens with random tokens that have the same POS tags and similar word embeddings. If the probability of classifying the perturbed text into the same class as that of the original text is above a predefined threshold, the selected predicates will be considered as the ultimate key features to interpret the prediction results. 

Similar to \citet{ribeiro2018anchors, ribeiro2016should}, \citet{alvarez2017causal} also proposed a model-agnostic interpretable method to relate inputs to outputs through the use of perturbed inputs generated by a variational auto-encoder applied to the original input. The perturbed input is supposed to have a similar meaning to the original input. A bipartite graph is then constructed to link these perturbed inputs and outputs, and the graph is then partitioned to highlight the relevant parts to show which inputs are relevant to the specific output tokens.

\citet{feng2018pathologies} proposed a method to gradually remove unimportant words from original texts while maintaining the model's performance. The remaining words are then considered as the important features for prediction. The importance of each token of the textual input is measured through a gradient approximation method, which involves taking the dot product between a given token's word embedding and the gradients of its output with respect to its word embedding \cite{ebrahimi2018hotflip}. The authors show that while the reduced inputs are nonsensical to humans, they are still enough for a given model to maintain a similar level of accuracy when compared with the original inputs.

The input perturbation method seems straightforward in identifying the significant input features by measuring the target task's performance changes with new perturbed instances. However, there are also works questioning the faithfulness of input perturbation. For example, \cite{slack2020fooling} conducted several experiments and argued that when the distributions of perturbed instances and original instances are less similar, the explanations of LIME \cite{ribeiro2016should} are not faithful. Another problem of most input perturbation explanations is that the identified important features are mostly independent tokens instead of coherent phrases like argued by \citet{ribeiro2018anchors}, which limits comprehensibility. The recent new track of local explanation: counterfactual explanations \cite{wu2021polyjuice, ribeiro2020beyond, chen2021kace} are generated via the approaches of input perturbation to provide counterfactual explanations to show what would happen if some certain features are replaced and prove those features are important for particular model decision. These counterfactual explanations extend beyond the input perturbation from the simple word-level to present the interpretation differently with the more straightforward counterfactual examples. Such presentation of the input perturbation interpretation would give normal users a more intuitive understanding.

\subsubsection{Attention weights}
Attention weight is a weighted sum score of input representation in intermediate layers of neural networks \cite{bahdanau2014neural}. Extracting attention weights for inputs to provide local interpretations for predictions is commonly used among models that utilise attention mechanisms. For NLP tasks with only textual inputs, tokens with higher attention weights are considered to have more impact on the outputs during the neural network training and are, therefore, regarded as the more important features. Attention weights have been used for explainability in sentiment analysis \citep{luo2018beyond,mao2019aspect,wang2018aspect}, question answering \citep{tu2020select,shen2018knowledge,sydorova2019interpretable}, and neural machine translation \citep{bahdanau2014neural,luong2015effective}. In tasks with both visual and textual inputs, such as Visual Question Answering (VQA)~\cite{lu2016hierarchical,yang2016stacked,yu2019deep,ding2023vqa,cao2023scenegate} and image captioning~\cite{xu2015show,anderson2018bottom,han-etal-2020-victr}, attention weights are extracted from both images and questions to identify the contributing features from both modalities. In the case of such multi-modal tasks, it is also important to boost the consistency between the attended image regions and sentence tokens for a plausible explanation. In recent years, different attention mechanisms have been proposed, including the self-attention mechanism \cite{vaswani2017attention} and the co-attention mechanism for multi-modal inputs \cite{yu2019deep}, aiming for better attention weights calculation that genuinely reflects the contributing factors to the final prediction. 

Though attention mechanisms have proved their effectiveness in performance increment in different tasks and have been used as the indicators of important features to explain the model's prediction results, there have always been debates arguing about the faithfulness of attention weights as the interpretation for neural networks.

\citet{bai2021attentions} proposed the concept of combinatorial shortcuts caused by the attention mechanism. It argued that the masks used to map the query and key matrices of the self-attention~\cite{vaswani2017attention} are biased, which would lead to the same positional tokens being attended regardless of the actual word semantics of different inputs. \citet{clark2019does} detected that the large amounts of attention of BERT \cite{bert} focus on the meaningless tokens such as the special token [SEP]. \citet{jain2019attention} argued that the tokens with high attention weights are not consistent with the important tokens identified by the other interpretable methods, such as the gradient-based measures. \citet{serrano2019attention} applied the method of intermediate representation erasure and claimed that attention can only indicate the importance of even intermediate components and are not faithful enough to explain the model's decision from the level of the actual inputs. 

In contrast, \citet{wiegreffe2019attention} proposed the work of `Attention is not not explanation' specifically against the arguments in \cite{jain2019attention}, arguing that whether attention weights are faithful explanations is dependent on the definition of explanation and conducted four different experiments to prove when attention can be used as the explanation. A similar view is also proposed by \citet{jacovi2020towards}, illustrating that under some instances, attention maps over input can be considered as a faithful explanation, which can be verified by the erasure method \cite{arras2017relevant, feng2018pathologies}, i.e. whether or not that erasing the attended tokens from inputs would change the prediction results.

In order to improve the faithfulness of attention as the explanation, some recent works have proposed different methods. For example, \citet{bai2021attentions} proposed a method of generating unbiased mask distribution by using random mask distributions to get attention weights through solely training the attention layers while fixing the other downstream parts of the model, which will therefore scale the attention weights towards tokens that are truly correlated with the predicted label. \citet{chrysostomou2021improving} introduced three different task-scaling mechanisms that scaled over the word representations from different aspects before passing to the attention mechanism and claimed that such scaled word representations help to produce a more faithful attention-based explanation. 

Overall, the dilemma of using inputs with high attention weights as the explanation to a black-box model's decision is associated with the various definition and inconsistent evaluations of explanation faithfulness from different works. \citet{jacovi2020towards} also proposed in their work that the possible approach to solving this issue is to construct a unified evaluation of the degree of faithfulness either from the level of a specific task or from the level of sub-spaces of the input space. Nevertheless, regardless of the debates over the faithfulness of attention, explanation by attention weights has a lower level of readability. Compared to rationale extraction works that explicitly force the consecutive rationales to be extracted for better comprehensibility, current works using attention as explanation neglect such interpretability aspect. Therefore, even in some cases where the input tokens with high attention weights could work as faithful explanations, it would be hard for non-experts to understand the explanation well with non-coherent highlighted tokens of the textual inputs. However, for the multimodal task such as the visual question answering, some works have attention maps over the images as the explanation \cite{luo2020rexup} or the part of the explanations \cite{wu-mooney-2019-faithful}, the attended region are usually consecutive pixels of the images, which can be more straightforward to be understood by non-expert users compared to the attention map over pure texts.

\subsubsection{Attribution Methods}
Another method of detecting important input features that contribute most to a specific prediction is attribution methods, which aim to interpret prediction outputs by examining the gradients of a model. Common attribution methods include DeepLift \cite{shrikumar2017learning}, Layer-wise relevance propagation (LRP) \cite{bach2015pixel}, deconvolutional networks \cite{zeiler2010deconvolutional} and guided back-propagation~\citep{springenberg2015striving}.

Extracting model gradients allows for identifying high-contributing input features to a given prediction. However, directly extracting gradients does not work well with regards to two key properties: sensitivity and implementation invariance. Sensitivity emphasizes that if we have two inputs with one differing feature that lead to different predictions, this differing feature should be noted as important to the prediction. Implementation invariance means that the outputs of two models should be equivalent if they are functionally equivalent, whether their implementations are the same or not. Focusing on these properties, \citet{sundararajan2017axiomatic} proposed an integrated gradient method. Integrated gradients are the accumulative gradients of all points on a straight line between an input and a baseline point (e.g. a zero-word embedding). \citet{he2019towards} applied this method to natural machine translation to find the contribution of each input word to each output word. Here, the baseline input is a sequence of zero embeddings in the same length as the input to be translated. \citet{mudrakarta2018did} applied integrated gradients to a question-answering task to identify the critical words in questions and found that only a few words in a question contribute to the model answer prediction. 

Besides extracting the gradients, scoring input contributions based on the model's hidden states is also used for attribution. For example, \citet{du2019attribution} proposed a post-hoc interpretable method that leaves the original training model untouched by examining the hidden states passed along by RNNs. \citet{ding2017visualizing} applied LRP \cite{bach2015pixel} to neural machine translation to provide interpretations using the hidden state values of each source and target word. 

The attribution methods are the preliminary approaches for deep learning researchers to explain the neural networks through the identified input features with outstanding gradients. The idea of the attribution methods were mostly proposed before the mature development and vast researches of rationale extraction, attention mechanisms and even the input perturbation methods. Compared to the other input feature explanation methods, the attribution methods hardly consider the interpretation's faithfulness and comprehensibility as the other three input feature explanation methods. Visualizing the identified features from inputs would be at the same plausible level as that of the other three feature importance methods to non-expert users, but the attribution methods do not work to form the interpretation into coherent sub-phrases for better readability and easier understanding. Thus, compared to rationale extraction, attention weights extraction and input perturbation, using attribution methods to generate the interpretation is more like a diagnosis method for deep learning experts to understand the model's decision and learn the model's functionality. 

\subsubsection{Datasets}

Tasks used for examining the interpretable methods discussed above include sentiment analysis, reading comprehension, natural machine translation, question answering and visual question answering. Below we list and summarise some common datasets that are used for these tasks:

\begin{enumerate}
    \item \textit{BeerAdvocate review dataset} \cite{mcauley2012learning} is a multi-aspect sentiment analysis dataset which contains around 1.5 million beer reviews written by online users. The average length of each review is about 145 words. These reviews are associated with the overall review of the beer or a particular aspect, such as the appearance, smell, palate and taste. Each written review also has a corresponding overall rating for beer and another four different ratings for the four review aspects, where each rating ranges from 0 to 5.
    
    \item \textit{IMDB} \cite{maas2011learning} is a large movie review usually used for binary sentiment classification. The dataset contains 50k reviews labelled as positive or negative and is split in half into train and test sets. The average length for each review is 231 words and 10.7 sentences. 
    
    \item \textit{WMT} is a workshop for natural machine translation. Tasks announced in these workshops include translation of different language pairs, such as French to English, German to English and Czech to English in WMT14, and Chinese to English additionally added in WMT17. The sources are normally news and biomedical publications. For many papers examining interpretable methods, the commonly used datasets are French to English news and Chinese to English news. 
    
    \item \textit{HotpotQA} \cite{yang2018hotpotqa} is a multi-hop QA dataset that contains 113K Wikipedia-based question-answer pairs where multiple documents are supposed to be used to answer each question. Apart from questions and answers, the dataset also contains sentence-level supporting facts for each document. This dataset is often used to experiment with interpretable methods for identifying sentence-level significant features for answer prediction.
    
    \item \textit{SQuAD} \cite{rajpurkar2016squad} is a reading comprehension dataset that contains 100k question-answer pairs from Wikipedia articles.  SQuAD v2 \cite{rajpurkar2018know} proposed in 2018 includes around 50K additional unanswerable questions to find the answerable questions with similar semantic meanings.
    
    \item \textit{VQA datasets} are used for multi-modal tasks with both textual and visual inputs. VQA v1 \cite{antol2015vqa} is the first visual question-answering dataset. VQA v1 contains 204,721 images, 614,163 questions and 7,964,119 answers, where most images are authentic images extracted from MS COCO dataset \cite{lin2014microsoft} and 50,000 images are newly generated abstract scenes of clipart objects. VQA v2 \cite{goyal2017making} is an improved version of VQA v1 that mitigates the biased question problem and contains 1M pairs of images and questions as well as ten answers for each question. Work on VQA commonly utilises attention weight extraction as a local interpretation method.
\end{enumerate}

\subsection{Natural Language Explanation}

Natural Language Explanation (NLE) refers to the method of generating free text explanations for a given pair of inputs and their prediction. In contrast to rational extraction, where the explanation text is limited to that found within the input, NLE is entirely freeform, making it an incredibly flexible explanation method. This has allowed it to be applied to tasks outside of NLP, including reinforcement learning \cite{ehsan2018rationalization}, self-driving cars \cite{kim2018textual}, and solving mathematical problems \cite{ling2017program}. We focus here on methods in which explanations are generated without any or minimal scaffolding, that is, we do not cover methods that form ‘natural language explanations’ by filling in templates, but rather cases where the explanation model is tasked with generating the entirety of the explanation content itself.

\subsubsection{Multimodal NLE}

Multimodal NLE focuses on generating natural language explanations for tasks that involve multiple input modalities, including images and video. While explanations may span multiple modalities, we focus on cases where the explanations significantly involve natural language. Much work, including text-only NLE, stems from \citet{hendricks2016generating}, which draws upon image captioning research to generate explanations for image classification predictions of bird images. The model first makes a prediction using an image classification network, and then the features from the final layers of the network are fed into an LSTM decoder \citep{hochreiter1997long} to generate the explanation text. The explanation is trained with a reinforcement learning-based approach both to match a ground truth correction and to be able to be used to predict the image label itself. Later work has directly built on this model by improving the use of image features used during the explanation generation \citep{Wickramanayake_Hsu_Lee_2019}, using a critic model to improve the relevance of the explanations \citep{hendricks2018generating}, and conditioning on specific image attributes \citep{8877393}. \citet{8579013} make use of an attention mechanism to augment the text-only explanations with heatmap-based explanations and find that training a model to provide both types of explanations improves the quality of both the text and visual-based explanations. Most of these earlier approaches use learnt LSTM decoders to generate the explanations, learning a language generation module from scratch. Most of these methods generate their explanations post-hoc, making a prediction before generating an explanation. This means that while the explanations may serve as valid reasons for the prediction, they may also not truthfully reflect the reasoning process of the model itself. \citet{wu-mooney-2019-faithful} attempt to build a multimodal model whose explanations better match the model’s reasoning process by training the text generator to generate explanations that can be traced back to objects used for prediction in the image as determined by gradient-based attribution methods. They explicitly evaluate their model’s faithfulness using LIME and human evaluation and find that this improves performance and does indeed result in explanations faithful to the gradient-based explanations.

More recently, NLE datasets have been developed for VQA \citep{huk2018multimodal}, self-driving car decisions \citep{kim2018textual}, arcade game agents \citep{10.1145/3301275.3302316}, visual commonsense \citep{zellers2019vcr}, physical commonsense \citep{rajani-etal-2020-esprit}, image manipulation detection \citep{da-etal-2021-edited}, explaining facial biometric scans \citep{9667024}, as well as for more general vision-language benchmarks \citep{Kayser_2021_ICCV}.

The recent rise of large pretrained language models \citep{peters-etal-2018-deep, devlin-etal-2019-bert, Radford2019LanguageMA} has also impacted multimodal NLE, with recent approaches replacing the standard LSTM-based decoder with pretrained text generation models such as GPT-2 \citep{marasovic-etal-2020-natural, Kayser_2021_ICCV, ayyubi2020generating} with a good deal of success. \citet{Kayser_2021_ICCV} additionally finds that using a pre-trained unified vision-language model along with GPT-2 works best over other combinations of vision and language-only models. This suggests that further utilising the growing number of large pre-trained multimodal models such as VLBERT \citep{Su2020VL-BERT}, UNITER \citep{chen2020uniter}, or MERLOT \citep{zellersluhessel2021merlot} may lead to improved explanations for multimodal tasks. However, while these models often do yield higher-quality explanations that better align with human preferences, the use of large unified transformer models means that the faithfulness of these explanations in representing the reasoning process of the model is hard to determine, as the exact reasoning processes used by these large models is hard to uncover.

\subsubsection{Text-only NLE}

Earlier work examining explanations accompanying NLP tasks largely examined integrating them as inputs for fact-checking, concept learning, and relation extraction \citep{srivastava-etal-2017-joint, hancock-etal-2018-training, alhindi-etal-2018-evidence}. These efforts provided useful datasets for examining natural language explanations, but the first work examining generating natural language explanations for NLP tasks in an automated fashion was done by \citet{esnli}, using a set of explanations gathered for the SNLI dataset \citep{bowman2015large} called e-SNLI. Similar to the multimodal models discussed above, the baseline models for e-SNLI proposed in \citet{esnli} are made up of two parts: a predictor module and an explanation module, with the best performing model first generating explanations and then using these explanations to make predictions. While this tighter integration of explanation generation into the overall model may suggest more faithful and higher-quality explanations, \citet{camburu-etal-2020-make} shows that this model can still provide explanations that are inconsistent with their predictions, suggesting that either the explanations are faulty or the model uses a flawed decision-making process. Several works try to improve the faithfulness of such models by using generated explanations as inputs to the final predictor model \citep{kumar-talukdar-2020-nile, zhao2021lirex, NEURIPS2020_4be2c8f2, rajani-etal-2019-explain}. By ‘explaining then predicting’, the explanations are by construction used as part of the prediction process. This may aid overall model performance by exposing latent aspects of the task \citep{hase-bansal-2022-models}. \citet{inoue-etal-2021-summarize} additionally shows summarisation models can be trained to serve as explanation generation models for this construction. However, recent work \citet{wiegreffe-etal-2021-measuring} suggests that jointly producing explanations actually results in models with a stronger correlation between the predicted label and explanation, suggesting these models are more faithful than explain-then-predict methods despite the different construction. Further evaluation linking the underlying model’s predictive mechanics with the generated explanations (e.g. \citet{prasad-etal-2021-extent} for highlighted rationales) may work to investigate further how much these explanations align with the underlying model.

Beyond NLI, other early tasks to which NLE was applied include commonsense QA \citep{rajani-etal-2019-explain} and user recommendations \citep{ni-etal-2019-justifying}. While early work used human-collected explanations, \citet{ni-etal-2019-justifying} shows that using distant supervision via rationales can also work well for training explanation-generating models. \citet{li-etal-2021-personalized} additionally embed extra non-text features (i.e., user id, item id) by using randomly initialised token embeddings. This provides a way to integrate non-text features besides the use of large pre-trained multimodal models.

Much like multimodal NLE, large pre-trained language models have also been integrated into text-based NLE tasks, and most recent papers make use of these models in some way. \citet{rajani-etal-2019-explain} introduce an NLE dataset for commonsense QA (`cos-e') and use a pre-trained GPT model \citep{Radford2018ImprovingLU} to generate explanations used to make a final prediction. More recently, wT5 \citep{narang2020wt5}, which follows the T5 model \citep{2020t5} in framing explanation generation and prediction as a purely text-to-text task, generates the prediction followed by a text explanation. More recent work has shown that using these models allows good explanation generation (and even may improve performance) for tasks and settings with little data \citep{10.1007/978-3-030-80599-9_8, marasovic-etal-2022-shot, DBLP:journals/corr/abs-2110-02056, yordanov2021fewshot}. Automatically collecting explanations from existing datasets or generating explanations using existing models can also provide extra supervision for learning to generate NLEs in limited-data settings \citep{Brahman_Shwartz_Rudinger_Choi_2021}. This highlights the strength of NLEs: by framing the explanation as a text generation problem, explanation generation is as simple as finetuning or even few-shot prompting a large language model to produce explanations, often with fairly good results. However, while these approaches are often impressive, generated explanations can still ‘hallucinate’ data not actually present in the training or input data and fail to generalise to challenging test sets such as HANS \citep{zhou-tan-2021-investigating}.

\subsubsection{NLE in Dialog}
While the above work has all assumed a setup where a model is able to generate only one explanation and has no memory of previous interactions with a user, some work has examined dialog-based setups where a user is assumed to repetitively interact with a model. \citet{DBLP:journals/corr/abs-1806-08055} propose a model for the components of an explanation dialog comprising of two sections: an explanation dialog, which consists mainly of presenting and accepting explanations; and an argument dialog, where the provided explanation is challenged with an argument. \citet{10.1145/3397481.3450676} draw on QA systems to design a model for explaining basic algorithms, presenting the model as an `interactive dialog that allows users to ask for specific kinds of explanations they deem useful'. More recently, \citet{li-etal-2022-using} use feedback from users as explanations to supervise and improve an open domain QA model, showing how models can improve by taking into account live feedback from users. Given the success of using human-written instructions to train large models \citep{sanh2022multitask, wei2022finetuned}, making further use of human feedback to improve and guide the way explanations are generated may further improve the quality and utility of NLEs.

\subsubsection{Datasets}

There are a number of NLE datasets for NLP tasks, which we summarise in Table \ref{tab:text_nle_dataset}. Many of these datasets consist of human-generated explanations applied to existing datasets, or make use of some automatic extraction method to retrieve explanations from supporting documents. While most datasets simply present one explanation per input sample, others present setups where multiple explanations are attached to each sample, but only one is valid \citep{wang-etal-2019-make, zhang-etal-2020-winowhy}. \citet{wiegreffe2021teach} also summarise existing NLE-for-NLP datasets, focussing also on text-based rationale and structured explanation datasets. We also provide a list of datasets for multimodal NLE in Table \ref{tab:modal_nle_dataset}.

\begin{table*}[]
\caption{Summary of datasets with natural language explanations for text-based tasks.}
\label{tab:text_nle_dataset}
\begin{tabular}{ccccc}
\toprule
\textbf{Ref.}                                                             & \textbf{Year} & \textbf{Dataset Name}                  & \textbf{Task}                                                                  & \textbf{Human-written explanations?}                                  \\
\midrule
\citep{esnli}                                   & 2016 & e-SNLI                        & NLI                                                                   & \cmark                                       \\
\citep{jansen-etal-2016-whats}                  & 2016 & - & Science Exam QA                                                       & \begin{tabular}[c]{@{}c@{}}Extracted from\\ auxiliary documents\end{tabular}                          \\
\citep{ling2017program}                         & 2017 & -             & Algebraic Word Problems                                               & \cmark                                       \\
\citep{srivastava-etal-2017-joint}              & 2017 & -   & Email Phsishing classification                                        & \cmark                                       \\
\citep{hancock-etal-2018-training}              & 2018 & BabbleLabble                  & Relation Extraction                                                   & \cmark                                       \\
\citep{alhindi-etal-2018-evidence}              & 2018 & LIAR-PLUS                     & Fact-checking                                                         & \begin{tabular}[c]{@{}c@{}}Extracted from\\ auxiliary documents\end{tabular}                 \\
\citep{rajani-etal-2019-explain}                & 2019 & cos-e                         & Commonsense QA                                                        & \cmark                                       \\
\citep{wang-etal-2019-make}                     & 2019 & -                    & Sense making                                                          & \cmark                                       \\
\citep{atkinson-etal-2019-gets}                 & 2019 & ChangeMyView                  & Opinion changing                                                      & \begin{tabular}[c]{@{}c@{}}Extracted from\\ reddit posts\end{tabular}                                  \\
\citep{zhang-etal-2020-winowhy}                 & 2020 & WinoWhy                       & Winograd Schema                                                       & \cmark                                       \\
\citep{kotonya-toni-2020-explainable-automated} & 2020 & PubHealth                     & Medical claim fact-checking                                           & \begin{tabular}[c]{@{}c@{}}Extracted from\\ auxiliary documents\end{tabular}                \\
\citep{Wang2020LearningFE}                      & 2020 & -  & \begin{tabular}[c]{@{}l@{}} Relation Extraction,\\ Sentiment Analysis \end{tabular}                        & \cmark                                       \\

\citep{Stammbach2020eFEVEREA}                   & 2020 & e-FEVER                       & Fact-checking                                                         & Generated using GPT-3    \\
\citep{aggarwal-etal-2021-explanations}         & 2021 & ECQA                          & Commonsense QA                                                        & \cmark                                       \\
\citep{Brahman_Shwartz_Rudinger_Choi_2021}  & 2021 & e-$\delta$-NLI                       & $\delta$-NLI  Rationale Generation & \begin{tabular}[c]{@{}c@{}}Extracted from\\ auxiliary documents,\\ automatically generated\end{tabular} \\
\bottomrule
\end{tabular}
\end{table*}

\begin{table*}[]
\caption{Summary of datasets with natural language explanations for multimodal tasks.}
\label{tab:modal_nle_dataset}
\begin{tabular}{ccccc}
\toprule\textbf{Ref.}                            & \textbf{Year} & \textbf{Dataset Name} & \textbf{Task}                                                                            & Human-written explanations? \\\midrule
\citep{huk2018multimodal}       & 2018 & VQA-X        & Visual QA                                                                       & \cmark                     \\
\citep{huk2018multimodal}       & 2018 & ACT-X        & Activity Recognition                                                            & \cmark                     \\
\citep{kim2018textual}          & 2018 & BDD-X        & \begin{tabular}[c]{@{}l@{}}Self-Driving Car\\ Decision Explanation\end{tabular} & \cmark                     \\
\citep{li2018vqae}              & 2018 & VQA-E        & Visual QA                                                                       & Generated from captions    \\
\citep{10.1145/3301275.3302316} & 2019 & -            & Frogger Game                                                                    & \cmark                     \\
\citep{zellers2019vcr}          & 2019 & VCR          & Visual Commonsense Reasoning                                                    & \cmark                     \\
\citep{rajani-etal-2020-esprit} & 2020 & ESPIRIT      & Physical Reasoning                                                              & \cmark                     \\
\citep{lei-etal-2020-likely}    & 2020 & VLEP         & Event Prediction                                                                & \cmark                     \\
\citep{da-etal-2021-edited}     & 2021 & EMU          & Understanding edits                                                             & \cmark    \\ 
\citep{Kayser_2021_ICCV}        & 2021 & E-ViL        & Vision-language Tasks                                                           & \cmark                     \\\bottomrule
\end{tabular}
\end{table*}

\subsubsection{Challenges and Future work}
NLE is very attractive as a human-comprehensible approach to interpretation: rather than trying to utilise model parameters, NLE-based approaches essentially allow their models to ‘talk for themselves’. Despite being freely generated, these explanations still display a degree of faithfulness in their agreement with gradient-based explanation methods and can be quite robust to noise \citep{wiegreffe-etal-2021-measuring}. This suggests that this approach exhibits a degree of faithfulness and stability despite a lack of formal guarantee that these methods have either quality. Furthermore, pipeline methods that use explanations for predictions can further guarantee that the generated explanations represent the information being used for prediction, even if their performance suffers compared to joint prediction models. NLEs have the benefit of being extremely comprehensible: unlike text rationales or gradient methods, which often require some understanding of the model being used, natural language explanations can be easily read and understood by anyone, and tailoring explanations to a specific audience is ‘simply’ a matter of training a model on similar explanations, which is even possible in low-data scenarios \citep{10.1007/978-3-030-80599-9_8, marasovic-etal-2022-shot, DBLP:journals/corr/abs-2110-02056, yordanov2021fewshot}. Finally, the trustworthiness of NLE methods is not often explicitly evaluated. The focus has been put on the overall ‘explanation quality’ when evaluating NLEs \citep{inoue-etal-2021-summarize, clinciu-etal-2021-study}. While rating ‘explanation quality’ may in some ways suggest how trustworthy the annotators find the explanations, more careful consideration of the type of contract-based trust \citep{10.1145/3442188.3445923} an NLE-based model may involve is required in determining the utility of deploying these models in real-world scenarios.

Overall, NLE is a very flexible and attractive explanation method, with the potential to greatly improve model explainability without requiring complex setups: just train your model to output explanations \citep{narang2020wt5}. However, evaluation must be carefully considered due to issues with automated metrics \citep{clinciu-etal-2021-study} and the human-generated explanations themselves \citep{carton-etal-2020-evaluating}. In addition, further exploring the link between generated NLEs and other explanation or interpretability methods may further yield insights into models and improve our understanding of the faithfulness of this method.

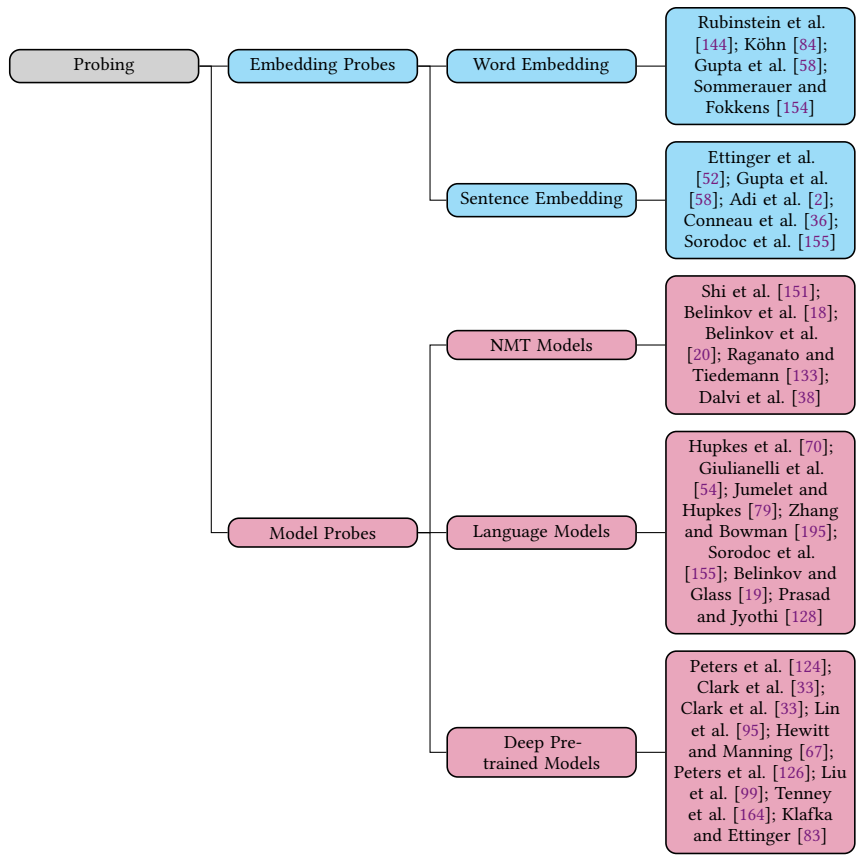
\begin{figure*}
\begin{forest}
  for tree={
    forked edges,
    font=\footnotesize,
    draw,
    grow' = 0,
    semithick,
    rounded corners,
    text width = 2.3cm,
    s sep = 6pt,
    node options = {align = center},
    calign=child edge, 
    calign child=(n_children()+1)/2
  }
  [Probing, fill=gray!35
    [Embedding Probes, for tree={fill=cyan!35}
        [Word Embedding
            [\citet{rubinstein-etal-2015-well};
            \citet{kohn-2015-whats};
            \citet{gupta-etal-2015-distributional};
            \citet{sommerauer-fokkens-2018-firearms}
            ]
        ]
        [Sentence Embedding
            [\citet{ettinger-etal-2016-probing};
            \citet{gupta-etal-2015-distributional};
            \citet{adi2016fine};
            \citet{conneau-etal-2018-cram};
            \citet{sorodoc-etal-2020-probing}
            ]
        ]
    ]
    [Model Probes, for tree={fill=purple!35}
        [NMT Models
            [\citet{shi-etal-2016-string};
            \citet{belinkov-etal-2017-neural};
            \citet{belinkov-etal-2017-evaluating};
            \citet{raganato-tiedemann-2018-analysis};
            \citet{dalvi2019one}
            ]
        ]
        [Language Models
            [\citet{hupkes2018visualisation};
            \citet{giulianelli-etal-2018-hood};
            \citet{jumelet-hupkes-2018-language};
            \citet{zhang-bowman-2018-language};
            \citet{sorodoc-etal-2020-probing};
            \citet{belinkov2019analysis};
            \citet{prasad-jyothi-2020-accents}
            ]
        ]
        [Deep Pre-trained Models
            [\citet{peters-etal-2018-dissecting};
            \citet{clark-etal-2019-bert};
            \citet{clark-etal-2019-bert};
            \citet{lin-etal-2019-open};
            \citet{hewitt-manning-2019-structural};
            \citet{peters2018deep};
            \citet{liu-etal-2019-linguistic};
            \citet{contextTenney};
            \citet{klafka-ettinger-2020-spying}
            ]
        ]
    ]
  ]
\end{forest}
\caption{Typology of Probing.}
\label{fig:typo-probing}
\end{figure*}

\subsection{Probing}

Linguistic probes, also referred to as `diagnostic classifiers' \citep{hupkes2018visualisation} or `auxiliary tasks' \citep{adi2016fine}, are a post-hoc method for examining the information stored within a model. Specifically, the probes themselves are (often small) classifiers that take as input some hidden representations (either intermediate representations within a model or word embeddings) and are trained to perform some small linguistic task, such as verb-subject agreement \citep{giulianelli-etal-2018-hood} or syntax parsing \citep{hewitt-manning-2019-structural}. Intuition follows that if there is more task-relevant information present within the hidden representations, the classifier will perform better, thus allowing researchers to determine the presence or lack of presence of linguistic knowledge within both word embeddings and at various layers within a model. However, recent research \citep{hewitt-manning-2019-structural, pimentel-etal-2020-information, ravichander-etal-2021-probing} has shown that probing experiments require careful design and consideration of truly faithful measurements of linguistic knowledge.

While current probing methods do not provide layperson-friendly explanations, they do allow for research into the behaviour of popular models, allowing a better understanding of what linguistic and semantic information is encoded within a model \citep{lin-etal-2019-open}. Hence, the target audience of a probe-based explanation is not a layperson, as is the case with other interpretation methods discussed in this paper, but rather an NLP researcher or ML practitioner who wishes to gain a deeper understanding of their model. Note we do not provide a list of common datasets in this section, unlike the previous sections, as probing research has largely not focused on any particular subset of datasets and can be applied to most text-based tasks.

\subsubsection{Embedding Probes}

Early work on probing focused on using classifiers to determine what information could be found in distributional word embeddings \citep{word2vec, glove}. For example, \citet{rubinstein-etal-2015-well, kohn-2015-whats, gupta-etal-2015-distributional} all investigated the information captured by word embedding algorithms through the use of simple classifiers (e.g. linear or logistic classifiers) to predict properties of the embedded words, such as part-of-speech or entity attributes (e.g. the colour of the entity referred to by a word). These works all found word embeddings captured the properties probed for, albeit to varying extents. More recently, \citet{sommerauer-fokkens-2018-firearms} used both a logistic classifier and a multi-layer perceptron (MLP) to determine the presence of certain semantic information in Word2Vec embeddings, finding that visual properties (e.g. colour) were not represented well, while functional properties (e.g. `is dangerous') were. Research into distributional models has reduced currently due to the rise of pre-trained language models such as BERT \citep{bert}.

Alongside word embeddings, sentence embeddings have also been the target of analysis via probing. \citet{ettinger-etal-2016-probing} (following \citet{gupta-etal-2015-distributional}) trains a logistic classifier to classify if a sentence embedding contains specific words and specific words with specific semantic roles. \citet{adi2016fine} trains MLP classifiers on sentence embeddings to determine if the embeddings contain information about sentence length, word content, and word order. They examine LSTM auto-encoder, continuous bag-of-words (CBOW), and skip-thought embeddings, finding that CBOW is surprisingly effective at encoding the properties of sentences examined in low dimensions, while the LSTM auto-encoder-based embeddings perform very well, especially with a larger number of dimensions. Further developing on this work, \citet{conneau-etal-2018-cram} proposes ten different probing tasks, covering semantic and syntactic properties of sentence embeddings and controlling for various cues that may allow a probe to `cheat' (e.g. lexical cues). In order to determine if encoding these properties aids models in downstream tasks, the authors also measure the correlation between probing task performance and performance on a set of downstream tasks. More recently, \citet{sorodoc-etal-2020-probing} proposes 14 additional new probing tasks for examining information stored in sentence embeddings relevant to relation extraction.

\subsubsection{Model Probes}

Following the work on probing distributional embeddings, \citet{shi-etal-2016-string} extended probing to NLP models, training a logistic classifier on the hidden states of LSTM-based neural machine translation (NMT) models to predict various syntactic labels. Similarly, they train various decoder models to generate a parse tree from the encodings provided by these models. By examining the performance of these probes on different hidden states, they find that lower-layer states contain more fine-grained word-level syntactic information, while higher-layer states contain more global and abstract information. Following this, \citet{belinkov-etal-2017-neural} and \citet{belinkov-etal-2017-evaluating} both examine NMT models with probes in more detail, uncovering various insights about the behaviour of NMT models, including a lack of powerful representations in the decoder, and that the target language of a model has little effect on the source language representation quality. Rather than a logistic classifier, both papers use a simple neural network with one hidden layer and a ReLU non-linearity, due to this reporting similar trends as a more simple classifier but with better performance. More recently, \citet{raganato-tiedemann-2018-analysis} analysed transformer-based NMT models using a similar probing technique alongside a host of other analyses. Finally, \citet{dalvi2019one} presented a method for extracting salient neurons from an NMT model by utilising a linear classifier, allowing examination of not just information present within a model but also what parts of the model contribute most to both specific tasks and the overall performance of the model.

Probing is not limited to NMT, however: research has also turned to examining the linguistic information encoded by language models. \citet{hupkes2018visualisation} utilised probing methods to explore how well an LSTM model for solving basic arithmetic expressions matches the intermediate results of various solution strategies, thus examining how LSTM models break up and solve problems with nested structures. Utilising the same method, \citet{giulianelli-etal-2018-hood} investigated how LSTM-based language models tracked agreement. The authors trained their probe (a linear model) on the outputs of an LSTM across timesteps and components of the model, showing how the information encoded by the LSTM model changes over time and in model parts. \citet{jumelet-hupkes-2018-language, zhang-bowman-2018-language} also probe LSTM-based models for particular linguistic knowledge, including NPI-licensing and CCG tagging. Importantly, the authors find that even untrained LSTM models contain information probe-based models can exploit to memorise labels for particular words, highlighting the need for careful control of probing tasks (we discuss this further in the next section). More recently, \citet{sorodoc-etal-2020-probing} probe LSTM and transformer-based language for referential information. We also note that probing has been applied to speech processing-based models \citep{belinkov2019analysis, prasad-jyothi-2020-accents}.

Finally, probing-based analyses of deep pre-trained language models have also been popular as a method for understanding how these models internally represent language. \citet{peters-etal-2018-dissecting} briefly utilised linear probes to investigate the presence of syntactic information in bidirectional LSTM models, finding that POS tagging is learnt in lower layers than constituent parsing. Recently, both \citet{lin-etal-2019-open} and \citet{clark-etal-2019-bert} used probing classifiers to investigate the information stored in BERT's hidden representations across both layers and heads. \citet{clark-etal-2019-bert} focused on attention, using a probe trained on attention weights in BERT to examine dependency information, while \citet{lin-etal-2019-open} focused on examining syntactic and positional information across layers. \citet{hewitt-manning-2019-structural} examined representations generated by ELMo \cite{peters2018deep} and BERT, training a small linear model to predict the distance between words in a parse tree of a given sentence. \citet{liu-etal-2019-linguistic} proposed and examined sixteen different probing tasks, involving tagging, segmentation, and pairwise relations, utilising a basic linear model. They compared results across several models, including BERT and ELMo, examining the performance of the models on each task across layers. \citet{contextTenney} trained two-layer MLP classifiers to predict labels for various NLP tasks (POS tagging, named entity labelling, semantic role labelling, etc.), using the representations generated by four different contextual encoder models. They found that the contextualised models improve more on syntactic tasks than semantic tasks when compared to non-contextual embeddings and found some evidence that ELMo does encode distant linguistic information. \citet{klafka-ettinger-2020-spying} investigated how much information about surrounding words can be found in contextualised word embeddings, training MLP classifiers to predict aspects of important words within the sentence, e.g. predicting the gender of a noun from an embedding associated with a verb in the same sentence.

\subsubsection{Probe Considerations and Limitations}

The continued growth of probing-based papers has also led to recent work examining best practices for probes and how to interpret their results. \citet{hewitt-liang-2019-designing} considered how to ensure that a probe is genuinely reflective of the underlying information present in a model and proposed the use of a control task, a randomised version of a probe task in which high performance is only possible by memorisation of inputs. Hence, a faithful probe should perform well on a probe task and poorly on a corresponding control task if the underlying model does indeed contain the information being probed for. The authors found that most probes (including linear classifiers) are over-parameterised, and discuss methods for constraining complex probes (e.g. multilayer perceptrons) to improve faithfulness while still allowing them to achieve similar results. 

While most papers we have discussed above follow the intuition that probes should avoid complex probes to prevent memorisation, \citet{pimentel-etal-2020-information} suggest that instead the probe with the best score on a given task should be chosen as the tightest estimate, since simpler models may simply be unable to extract the linguistic information present in a model, and such linguistic information cannot be `added' by more complex probes (since their only input are hidden representations). In addition, the authors argue that memorisation is an important part of linguistic competence, and as such probes should not be artificially punished (via control tasks) for doing this. Recent work has also presented methods that avoid making assumptions about probe complexity, such as MDL probing \citep{voita-titov-2020-information, lovering2021predicting}, which directly measures ‘amount of effort’ needed to achieve some extraction task,  or DirectProbe \citep{zhou-srikumar-2021-directprobe}, which directly examines intermediate representations of models to avoid having to deal with additional classifiers.

Finally, \citet{hall-maudslay-etal-2020-tale} compared the structural probe \cite{hewitt-manning-2019-structural} with a lightweight dependency parser (both given the same inputs) and demonstrated that the parser is generally able to extract more syntactic information from BERT embedding. In contrast, the probe performs better with a different metric, showing that the choice of metric is important for probes: when testing for evidence of linguistic information, one should consider not only the nature of the probe but also the metric used to evaluate it. Furthermore, the significance of well-performing probes is not clear: models may encode linguistic information not actually used by the end-task \citep{ravichander-etal-2021-probing}, showing that the presence of linguistic information does not imply it is being used for prediction. Some approaches proposed later that integrated the causal approaches such as amnesiac probing \citep{10.1162/tacl_a_00359}, which directly intervene in the underlying model’s representations, might be a possible solution to distinguish between these cases.

\subsubsection{Interpretability of Probes and Future Work}
As noted at the beginning of the section, probing is a way for NLP researchers to investigate models rather than end-users. As such, their comprehensibility is relatively low: understanding probing results requires understanding the linguistic properties they are probing and the more complex experimental setups they make use of (as simple metrics such as task accuracy do not show the whole story \citep{hewitt-liang-2019-designing}). However, probes are naturally fairly faithful in that they directly use the model's hidden states and are specifically designed to represent only information present within these hidden states. This faithfulness is degraded somewhat by the fact that this information may not be used for predictions \citep{ravichander-etal-2021-probing}, but recent causal approaches work towards alleviating this. This also suggests that probing results could be considered trustworthy only when the experimental design is carefully considered, in that their results can only be relied upon if carefully controlled. Finally, probing methods are often reasonably stable for the same model and property, as the probe classifier is trained to some convergence. However, across models (even those with the same architecture but just trained on different data), results can differ quite drastically \citep{10.1162/tacl_a_00359}, which shows differences between pre-trained and finetuned BERT models. This is more likely to be a function of the underlying models rather than the technique, but also shows that probing results are very specific to the models and properties being examined.

Overall, probes are exciting and valuable tools for investigating models' ‘inner workings’. However, much like other explanation methods, the setup and evaluation of probing techniques must be carefully considered. Some future works of the probing may be associated with the integration of some causal methods \citep{10.1162/tacl_a_00359} as better approaches to make stronger statements about what a model is and isn’t using for its predictions, better allowing probing to provide explanations for model judgements rather than just show what could be potentially used. Combining this with methods that further reduce the complexity of probing setups \citep{zhou-srikumar-2021-directprobe} may allow even simpler and better ways to get insights into NLP models. Causal models have been applied to the traditional predictive tasks and covered with the convergence of causal inference and language processing \cite{feder2022causal}. Recent NLP works have tried to involve auxiliary causal-based approaches in their models \cite{feder2022causal,heskes2020causal}. Such involvement of causal approaches can be seen as a future trend of interpretable NLP tasks including probing. However, the essence of causal models is different from the association essence of neural networks. Thus, we consider a detailed discussion of causal approaches is out of the scope of this survey. But we do notice that this could be a future trend for the further development of probing.

\section{Evaluation Methods}

\subsection{Evaluation of Feature Importance}

\subsubsection{Automatic Evaluation} 
Evaluations on the interpretable methods of extracting important features usually align with the evaluation of the explanation faithfulness, i.e. whether the extracted features are sufficient and accurate enough to result in the same label prediction as the original inputs. When the datasets come with pre-annotated explanations, the extracted features used as the explanation can be compared with the ground truth annotation through exact matching or soft matching. The exact matching only considers the validness of the explanation when it is exactly the same as the annotation, and such validity is quantified through the precision score. For example, the HotpotQA dataset provides annotations for supporting facts, allowing a model's accuracy in reporting these supporting facts to be easily measured. This is commonly used for extracting rationals, where the higher the precision score, the better the model matches human-annotated explanations, likely indicating improved interpretability. On the country, soft matching will take the extracted features as a valid explanation if some features (tokens/phrases in the case of NLP) matched with the annotation. For instance, \citet{deyoung2020eraser} proposed Intersection-Over-Union (IOU) on the token level, taking the overlap size of the tokens over two spans divided by the union of their token sizes and considering the extracted rationales as a valid explanation if the IOU score is over 0.5.

However, \citet{deyoung2020eraser} also argued that the matching between the identified features and the annotation only measures the plausibility of interpretability but not faithfulness. In other words, either the exact matching or soft matching can reveal if the model's decisions truly depend on the identified contributing features. Therefore, some other erasure-based metrics are specifically proposed to evaluate the impact of the identified important features to the model's results. For example,~\citet{du2019attribution} proposed a \textbf{faithfulness score} to verify the importance of the identified contributing sentences or words to a given model's outputs. It is assumed that the probability values for the predicted class will significantly drop if the truly important inputs are removed. The score is calculated as in equation~\ref{faithfulness}:
\begin{equation}\label{faithfulness}
    S_{Faithfulness} = \frac{1}{N}\sum_{i=1}^{N}\left ( y_{x^{i}}-y_{x_{\setminus A}^{i}} \right )
\end{equation}
where $y_{x^{i}}$ is the predicted probability for a given target class with original inputs and $y_{x_{\setminus A}^{i}}$ is the predicted probability for the target class for the input with significant sentences/words removed.

The \textbf{Comprehensiveness score} proposed by \citet{deyoung2020eraser} in later years is calculated in the same way as the Faithfulness score~\cite{du2019attribution}. What is to be noted here is that the Comprehensiveness score is not related to the evaluation of the comprehensibility of interpretability but to measure whether all the identified important features are needed to make the same prediction results. A high score implies the enormous influence of the identified features, while a negative score indicates that the model is more confident in its decision without the identified rationales. \citet{deyoung2020eraser} also proposed a \textbf{Sufficiency score} to calculate the probability difference from the model for the same class once only the identified significant features are kept as the inputs. Thus, opposite to the Comprehensiveness score or Faithfulness score, a lower Sufficiency score indicates the higher faithfulness of the selected features. 

Apart from using the above-proposed evaluation metrics, another direct way to evaluate the validity of the explanations for a model's output is to examine the \textbf{performance decrease} of a model based on the tasks standard performance evaluation metrics after removing or perturbing identified important input features (i.e. words/phrases/sentences). For example, ~\citet{he2019towards} measured the change in BLEU scores to examine whether certain input words were essential to the predictions in natural machine translation.

\subsubsection{Human Evaluation}
Human evaluation is also a common and straightforward but relatively more subjective method for evaluating the validity of explanations for a model. This can be done by researchers themselves or by a large number of crowd-sourced participants (sourced from, e.g. Amazon Mechanical Turk). For example, \citet{chen2018learning} asked Amazon Mechanical Turk workers to predict the sentiment based on predicted keywords in a text, examining the faithfulness of the selected features as interpretation. \citet{sha2021learning} sampled 300 input-output-interpretation cases to ask the human evaluator to examine whether the selected features are useful (to explain the output), complete (enough to explain the output) and fluent to read). 

While faithfulness can be evaluated more easily via automatic evaluation metrics, the comprehensibility and trustworthiness of interpretations usually are evaluated through human evaluations in the current research works. Though using large numbers of participants helps remove the subjective bias, this requires the cost of setting up larger-scale experiments, and it is also hard to ensure that every participant understands the task and the evaluation criteria. It is undoubtedly that the human evaluation results can provide some hints about the interpretation validity and comprehensibility, but we cannot erase the suspicion of the existence of subjective bias, which also limits further references and fair comparison of the human evaluation results for future works.

\subsection{Evaluation of NLE}

\subsubsection{Automatic Evaluation}

As NLE involves generating text, the automatic evaluation metrics for NLE are generally the same metrics used in tasks with free-form text generation, such as machine translation or summarization. As such, standard automated metrics for NLE are BLEU \citep{papineni-etal-2002-bleu}, METEOR \citep{denkowski:lavie:meteor-wmt:2014}, ROUGE \cite{lin-2004-rouge}, CIDEr \citep{vedantam2015cider}, and SPICE \citep{anderson2016spice}, with all five generally being reported in VQA-based NLE papers. Perplexity is also occasionally reported \citep{esnli, ling2017program}, keeping in line with other natural language generation-based works. However, these automated metrics must be used carefully, as recent work has found they often correlate poorly with human judgements of explanation quality. \citet{clinciu-etal-2021-study} suggest that model-based scores such as BLUERT and BERTScore better correlate with human judgements, and \citet{hase-etal-2020-leakage} point out that only examining how well the explanation output matches labels does not measure how well the explanations accurately reflect the model’s behaviour. 

Additionally, the quality of the annotated human explanations collected in datasets such as e-SNLI has also come into question. \citet{carton-etal-2020-evaluating} find that human-created explanations across several datasets perform poorly at metrics such as sufficiency and comprehensiveness, suggesting they do not contain all that is needed to explain a given judgement. This suggests that just improving our ability to compare generated explanations with human-generated ones may not be enough to best measure the quality of a given generated explanation, and further work in improving the gold annotations provided by explanation datasets could also help.

\subsubsection{Human Evaluation}

Given the limitations of current automatic evaluation methods, and the free-form nature of NLE, human evaluation is always necessary to truly judge explanation quality. Such evaluation is most commonly done by getting crowdsourced workers to rate the generated explanations (either just as correct/not correct or on a point scale), which allows easy comparison between models. In addition, \citet{liu-etal-2019-towards-explainable} uses crowdsourced workers to compare their model's explanations against another, with workers noting which model's explanation related best to the final classification results. Considering BLEU and similar metrics do not necessarily correlate well with human intuition, all work on NLE should include human evaluation results to some level, even if the evaluation is limited (e.g. just on a sample of generated explanations).

\subsection{Evaluation of Probing}

As probing tasks are more tests for the presence of linguistic knowledge rather than explanations, the evaluation of probing tasks differs according to the tasks. However, careful consideration should be given to the choice of metric. As \citet{hall-maudslay-etal-2020-tale} showed, different evaluation metrics can result in different apparent performances for different methods, so the motivation behind a particular metric should be considered. Beyond metrics, \citet{hewitt-liang-2019-designing} suggested that the \textit{selectivity} of probes should also be considered, where selectivity is defined as the difference between probe task accuracy and control task\footnote{A control task being a variant of the probe task which utilises random outputs to ensure that high scores on the task are only possible through `memorisation' by the probe.} accuracy. While best practices for probes are still being actively discussed in the community \cite{pimentel-etal-2020-information}, control tasks are undoubtedly helpful tools for further investigating and validating the behaviour of models uncovered by probes.

\section{Discussion and Conclusion}
This paper focused on the local interpretable methods commonly used for natural language processing models. In this survey, we have divided these methods into three different categories based on their underlying characteristics: 1) explaining the model's outputs from the input features, where these features could be identified through rationale extraction, perturbing inputs, traditional attribution methods, and attention weight extraction; 2) generating the natural language explanations corresponding to each input; 3) using diagnostic classifiers to analyse the hidden information stored within a model. For each method type, we have also outlined the standard datasets used for different NLP tasks and different evaluation methods for examining the validity and efficacy of the explanations provided. 

By going through the current local interpretable methods in the field of NLP, we identified several limitations and research gaps to be overcome to develop explanations that can stably and faithfully explain the model's decisions and be easily understood and trusted by users. Firstly, as we have stated in section \ref{sec: Explainability vs Interpretability}, there is currently no unified definition of interpretability across the interpretable method works. While some researchers distinguish interpretability and explainability as two separate concepts \cite{rudin2018please} with different difficulty levels, many works use them as synonyms of each other, and our work also follows this way to include diverse works. However, such an ambiguous definition of interpretability/explainability leads to inconsistent interpretation validity for the same interpretable method. For example, the debate about whether the attention weights can be used as a valid interpretation/explanation between \citet{wiegreffe2019attention} and \citet{jain2019attention} is due to the conflicting definition. The argument of \citet{jain2019attention} is based on the fact that only the faithful interpretable methods are truly interpretable, while \citet{wiegreffe2019attention} argued that attention is an explanation if we accept that explanation should be plausible but not necessarily faithful as proposed by \cite{rudin2018please}. Thus, we need a unified and legible definition of interpretability that should be broadly acknowledged and agreed to help further develop valid interpretable methods.  

Secondly, we need effective evaluation methods that can evaluate the multiple dimensions of interpretability, the results of which can be reliable for future baseline comparison. However, the existing evaluation metrics measure only limited interpretability dimensions. Taking the evaluation of rationales as an example, examining the matching between the extracted rationales and the human rationales only evaluates the plausibility but not faithfulness \cite{deyoung2020eraser}. However, when it comes to the faithfulness evaluation metrics \cite{chrysostomou2021improving, serrano2019attention, deyoung2020eraser, arya2019one}, the evaluation results on the same dataset can be opposite by using different evaluation metrics. For example, two evaluation metrics DFFOT \cite{serrano2019attention} and SUFF \cite{deyoung2020eraser} concludes opposite evaluation results on LIME method of the same dataset \cite{chan2022comparative}. Moreover, the current automatic evaluation approaches mainly focus on the faithfulness and comprehensibility of interpretation. It can hardly be applied to evaluate the other dimensions, such as stability and trustworthy. The evaluation of other interpretability dimensions relies too much on the human evaluation process. Though human evaluation is currently the best approach to evaluate the generated interpretation from various aspects but can be subjective and less reproducible. In addition, it is also essential to have efficient evaluation methods that can evaluate the validity of interpretation in different formats. For example, the evaluation of the faithful NLE relies on the BLEU scores to check the similarity of generated explanations with the ground truth explanations. However, such evaluation methods neglect that the natural language explanations with different contents from the ground truth explanations can also be faithful and plausible for the same input and output pair. To sum up, there is still a considerable research gap for developing effective evaluation methods and frameworks to verify the interpretable methods from various dimensions. and such development would also require explainable datasets with good-quality annotations. The evaluation framework should provide fair results that can be reused and compared by future works, and should be user-centric, taking into account the aspects of different groups of users \cite{kaur2020interpreting}.

\section{Future trend of Interpretability}
The future trend of developing interpretable methods cannot avoid further conquering the current limitations. Developing truly faithful interpretable methods that can precisely explain the model's decisions is critical to enable the vast application of deep neural networks to crucial fields, including medicine, justice and finance. Faithful interpretable methods and easily understandable interpretations are key to bringing users' trust to the model's decisions, especially for users without deep learning knowledge. It is natural for them to question the decisions from an unfamiliar technique. Providing faithful, comprehensible and stable interpretations of a model helps eliminate the questions and uncertainties about using a black-box model for any users. 

However, apart from the discussed limitations of the current interpretable methods, one existing problem is that evaluating whether an interpretation is faithful mainly considers the interpretations for the model's correct predictions. In other words, most existing interpretable works only explain why an instance is correctly predicted but do not give any explanations about why an instance is wrongly predicted. If the explanations of a model's correct predictions precisely reflect the model's decision-making process, this interpretable method will usually be regarded as a faithful interpretable method. However, it is also significant and irradiative to generate explanations of the wrong prediction results to investigate and examine which parts of the input instances the model attended to when made the wrong decision and whether those parts can reflect the model's wrong decision-making process. However, the interpretation and explanation of the model's wrong prediction are not considered in any existing interpretable works. Some works even directly consider the interpretations generated by their interpretable models for the models' wrong predictions are invalid and incorrect~\cite{8579013, marasovic-etal-2020-natural, wu-mooney-2019-faithful, Kayser_2021_ICCV} and therefore, would not being taken into account for the measurement of intepretbality faithfulness. This seems reasonable when the current works are still struggling with developing interpretable methods that can at least faithfully explain the model's correct predictions. However, the interpretation of a model's decision should not only be applied to one side but to both correct and wrong prediction results. 

This also brings us the reflection that the fundamental reason to develop model interpretability is more than providing evidence/support/explanation of a correct prediction to users to make them believe the model's correct decisions, but also to give them valuable guidance about why the model makes a wrong prediction. The comprehensive interpretations of a model's decisions should provide faithful explanations for the model's both correct and incorrect predictions. Such comprehensive interpretations from both sides are the key to developing the ultimate trustworthiness for black-box models and boosting their broader and more stable applications in required fields. Moreover, understanding the reason for the wrong prediction is also essential for deep learning researchers to learn and adjust the model better in the future works.

Therefore, the future works of interpretability would be to fill the current research gap and develop interpretable models that can generate faithful and comprehensible interpretations for both correct and incorrect decisions made by the model, providing reliable information to improve the trust of non-experts in using deep neural networks in crucial fields and help experts understand and improve the model in a more accurate and better way.

\bibliographystyle{ACM-Reference-Format}
\bibliography{ref}


\end{document}